\documentclass[10pt,twocolumn,letterpaper]{article}

\usepackage{iccv}
\usepackage{times}
\usepackage{epsfig}
\usepackage{graphicx}
\usepackage{amsmath}
\usepackage{amssymb}

\usepackage{url}            
\usepackage{booktabs}       
\usepackage{amsfonts}       
\usepackage{nicefrac}       
\usepackage{microtype}      
\usepackage[ruled,lined]{algorithm2e}
\usepackage{color}
\usepackage{float}
\usepackage{bbm}
\usepackage{multirow}
\usepackage{capt-of}
\usepackage{color, colortbl}

\usepackage[colorlinks=true,breaklinks=true,bookmarks=false]{hyperref}

\iccvfinalcopy 

\def\iccvPaperID{0890} 

\ificcvfinal\fi

\begin{document}

\title{Dynamic Kernel Distillation for Efficient Pose Estimation in Videos}
\author{\normalsize{Xuecheng~Nie$^{1,}$\thanks{This work was partly done while Xuecheng was an intern as Snap Inc.}} \quad  \qquad \normalsize{Yuncheng~Li$^{2}$} \quad  \qquad \normalsize{Linjie~Luo$^3$} \quad  \qquad \normalsize{Ning~Zhang} \qquad \quad  \normalsize{Jiashi~Feng$^1$}\\
	\small{$^{1}$Department of Electrical and Computer Engineering, National University of Singapore, Singapore} \\
	\small{$^{2}$Snap Inc. \quad $^{3}$ByteDance AI Lab} \\
	{\small \tt niexuecheng@u.nus.edu}  \ \ {\small\tt raingomm@gmail.com} \ \ {\small\tt linjie.luo@bytedance.com} \\ {\small\tt ningzhang@berkeley.edu} \ \  {\small \tt elefjia@nus.edu.sg}
}

\maketitle
\ificcvfinal\thispagestyle{empty}\fi

\begin{abstract}
Existing video-based human pose estimation methods extensively apply large networks onto every frame in the video to localize body joints, which  suffer high computational cost and hardly meet the low-latency requirement in realistic applications. To address this issue, we propose a novel Dynamic Kernel Distillation (DKD) model to facilitate small networks for estimating human poses in videos, thus significantly lifting the efficiency. In particular, DKD introduces a light-weight distillator to online distill pose kernels via leveraging temporal cues from the previous frame in a one-shot feed-forward manner. Then, DKD simplifies body joint localization into a matching procedure between the pose kernels and the current frame, which can be efficiently computed via  simple convolution. In this way, DKD fast transfers pose knowledge from one frame to provide compact guidance for body joint localization in the following frame, which enables utilization of small networks in video-based pose estimation. To facilitate the training process, DKD exploits a temporally adversarial training strategy that introduces  a temporal discriminator to help generate temporally coherent pose kernels and pose estimation results  within a long range. Experiments on Penn Action and Sub-JHMDB benchmarks demonstrate outperforming efficiency of DKD, specifically, $10{\times}$ flops reduction and  $2{\times}$ speedup over previous best model, and its state-of-the-art accuracy.
\end{abstract}

\section{Introduction}

\begin{figure}[t!]
	\begin{center}
		\includegraphics[scale=0.85]{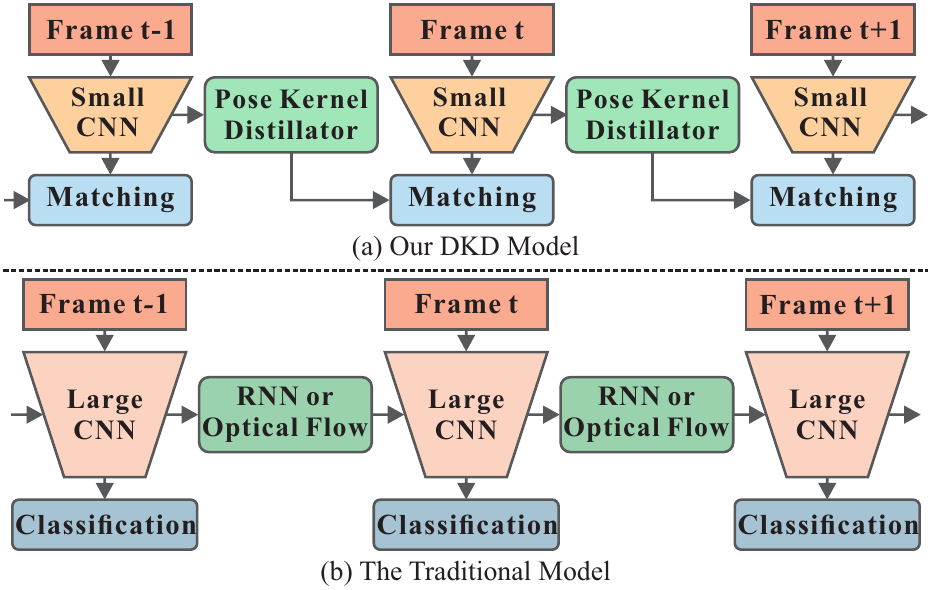}
		\caption{Comparison between (a) our DKD model and (b) the traditional model for video-based human pose estimation. DKD online distills coherent pose knowledge and simplifies body joint localization into a matching procedure, facilitating small networks to efficiently estimate human pose in videos while achieving outperforming accuracy. See text for details.}
		\label{fig:fig1}
	\end{center}
	\vspace{-8mm}
\end{figure}

Human pose estimation in videos aims to generate framewise joint localization of the human body. It is important for many applications  including surveillance~\cite{cristani2013human}, computer animation~\cite{lee2002interactive}, 
and AR/VR~\cite{lin2010augmented}. Compared to its still-image based counterpart, this task is more challenging due to its low-latency requirement and various  distracting factors, \emph{e.g.}, motion blur, pose variation and viewpoint change.

Prior CNN based methods to solve this task~\cite{girdhar2018detect,xiaohan2015joint,song2017thin,luo2017lstm} usually use a large network to extract representative features for every frame and localize body joints based on them via pixel-wise classification. Some recent works also incorporate temporal cues from optical flow~\cite{dosovitskiy2015flownet} or RNN units~\cite{xingjian2015convolutional} to improve the performance, as shown in Fig.~\ref{fig:fig1}~(b). Despite their notable accuracy, these methods suffer  expensive computation cost from the large model size, and hardly meet the low-latency requirement for realistic applications. The {efficiency} of video-based pose estimation still needs to be largely enhanced. 

In this paper, we propose to  enhance  efficiency of human pose estimation in videos by fully leveraging temporal cues to enable small networks to   localize body joints accurately.
Such an idea is motivated by observing the computational bottleneck for prior models. Considering the temporal consistency across adjacent frames, it is not necessary to pass every frame through a large network for feature extraction. Instead, the model only needs to learn how to effectively transfer  knowledge of   pose localization in previous frames to the subsequent frames. Such transfer can help alleviate the requirements of large models and reduce the overall computational cost.

To implement the above idea, we design a novel Dynamic Kernel Distillation (DKD) model. As shown in Fig.~\ref{fig:fig1}~(a), DKD online distills pose knowledge from   the previous frame into pose kernels through a light-weight distillator. Then, DKD simplifies body joint localization into a matching procedure between the pose kernels and the current frame through simple convolution. In this way, DKD fast re-uses pose knowledge from one frame and provides compact guidance for a small network to learn discriminative features for accurate human pose estimation. 

In particular, DKD introduces a light-weight CNN based pose kernel distillator. It takes features and pose estimations of the previous frame as input and infers   pose kernels suitable for the current frame. These pose kernels carry knowledge of body joint configuration patterns from the previous frame to the current frame, and guide  a small network to learn compact features matchable to the pose kernels  for efficient pose estimation.
Accordingly, body joint localization is cast as a matching procedure via applying pose kernels on  
feature maps output from small networks
with simple convolution to search for regions with similar patterns. Since it gets rid of the need for using large networks, DKD performs significantly faster than prior models. In addition, this 2D convolution based matching scheme is significantly cheaper than additional optical flow~\cite{dosovitskiy2015flownet}, the decoding phase of an RNN unit~\cite{xingjian2015convolutional} or the expensive 3D convolutions~\cite{ji20133d}. 
Moreover, the distillator framewisely updates the pose kernels according to current joint representations and configurations. This dynamic feature makes DKD more flexible and robust in analyzing various scenarios in videos.

To further leverage temporal cues to  facilitate the  distillator to infer suitable pose kernels, DKD introduces a temporally adversarial training method that adopts a discriminator to help estimate consistent poses in consecutive frames. The temporally adversarial discriminator learns to distinguish the groundtruth change of  joint confidence maps over neighboring frames from the predicted change, and thus supervises DKD to generate temporally coherent poses. In contrast to previous adversarial training methods~\cite{chou2017self,chen2017adversarial} that learn structure priors in the spatial dimension for recognition over still images, our  method  constrains the pose variations in the temporal dimension of videos, enforcing plausible changes of estimated poses in videos. In addition, this discriminator can be removed during the inference phase, thus introducing no additional computation.

The whole framework of the proposed DKD model is end-to-end learnable. Comprehensive experiments on two widely used benchmarks Penn Action~\cite{zhang2013actemes} and Sub-JHMDB~\cite{jhuang2013towards} demonstrate the efficiency and effectiveness of our DKD model for resolving human pose estimation in videos. Our main contributions are in three folds: 
1) We propose a novel model to facilitate small networks 
in video-based pose estimation with lifted efficiency,
by using a light-weight distillator to online distill the pose knowledge and simplifying body joint localization into a matching procedure with simple convolution.
2) We introduce the first temporally adversarial training strategy for encouraging the coherence of estimated poses in the temporal dimension of videos. 
3) Our model achieves outperforming efficiency, \emph{i.e.} 10x flops reduction and 2x  speedup over previous best model, also with state-of-the-art accuracy.

\section{Related work}

For human pose estimation in videos,
existing CNN based methods~\cite{grinciunaite2016human,gkioxari2016chained,song2017thin,luo2017lstm,girdhar2018detect}
usually focus on leveraging temporal cues to extract complementary information for refining the preliminary results output from a large network for every frame. In~\cite{iqbal2017pose}, Iqbal~\emph{et al.} incorporate deep learned representations into an action conditioned pictorial structured model to refine pose estimation results of each frame. 
In~\cite{grinciunaite2016human} and~\cite{girdhar2018detect}, 3D convolutions are exploited on video clips for implicitly capturing the temporal contexts between frames. 
In~\cite{song2017thin}, Song~\emph{et al.} propose a Thin-Slicing network that uses dense optical flow to warp and align heatmaps of neighboring frames and then performs spatial-temporal inference via message passing through the graph constructed by joint candidates and their relationships among aligned heatmaps.
\cite{gkioxari2016chained} and~\cite{luo2017lstm} sequentially estimate human poses in videos following the Encoder-RNN-Decoder framework. Given a frame, this kind of framework first uses  an encoder network to learn high-level image representations, then RNN units to explicitly propagate temporal information between neighboring frames and produce hidden states, and finally a decoder network to take hidden states as input and output pose estimation results of current frame. For ensuring good performance, however, these methods always require large network to compactly learn intermediate representations or preliminary poses. Their efficiency is rather limited.

Different from existing methods, our DKD model distills coherent pose knowledge from temporal cues and simplifies body joint localization as a matching problem, thus allowing small networks to accurately and efficiently estimate human poses in videos, which is explained in more detail below.   

\begin{figure*}[t!]
	\begin{center}
		\includegraphics[scale=0.68]{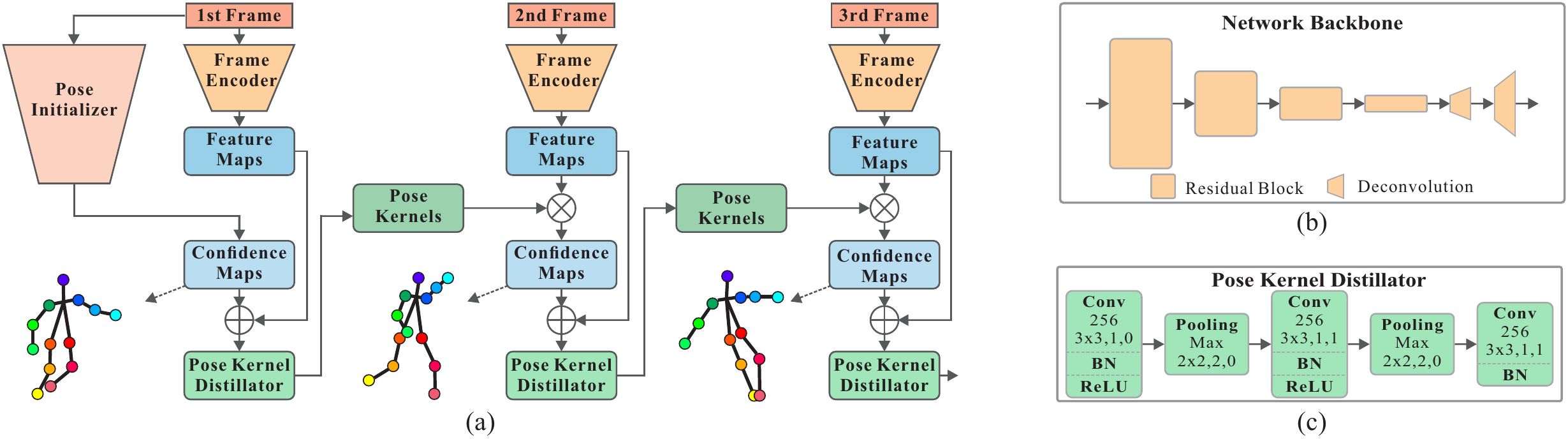}
		\caption{The architecture of the proposed Dynamic Kernel Distillation model. (a) The overall framework of the DKD model for inferencing human poses in videos. $\otimes$ denotes the convolution operation and $\oplus$ the concatenation. (b) The network backbone utilized in the pose initializer and frame encoder. (c) The network architecture of the pose kernel distillator. 
		}
		\label{fig:flow_chart}
	\end{center}
	\vspace{-8mm}
\end{figure*}

\section{Proposed approach}

\subsection{Formulation}
We first mathematically formulate the proposed Dynamic Kernel Distillation (DKD) model for human pose estimation in videos. For a video $\mathcal{V}{=}\{I_t\}_{t=1}^T$ including $T$ frames, we use $I_t{\in}\mathbb{R}^{M{\times}N{\times}3}$ to denote its $t$th frame, where $M$ and $N$ are the height and width of $I_t$, respectively. DKD aims to estimate a set of confidence maps $\mathcal{H}{=}\{h_t\}_{t=1}^T$ for all frames in $\mathcal{V}$. The $h_t{\in}\mathbb{R}^{m{\times}n{\times}K}$ is of spatial size $m{\times}n$, where $K$ is the number of body joints, and each of its elements encodes the confidence of a joint at the corresponding position. Accordingly, DKD performs online human pose estimation frame-by-frame in a sequential manner, by leveraging temporal cues between neighboring frames. In particular, its core is composed of a pose kernel distillator with a temporally adversarial training strategy.

\vspace{-4mm}
\paragraph{Pose kernel distillation} 
Given a frame $I_t$, DKD introduces a pose kernel distillator $\Phi(\cdot)$ to transfer pose knowledge provided by $I_t$ to guide pose estimation in the next frame $I_{t+1}$. 
In particular, it leverages temporal cues represented with the combination of feature maps $f_t$ and confidence maps $h_t$, to \emph{online} distill pose kernels $k_t$ via a simple feed-forward computation
\begin{equation}
k_t = \Phi(f_t, h_t),
\end{equation}
where $k_t{\in}\mathbb{R}^{S{\times}S{\times}C{\times}K}$ and $S$ is the kernel size. The distilled pose kernels $k_t$ encode  knowledge of body joint patterns and provide compact guidance for pose estimation in the posterior frame, which is learnable with light-weight networks. Accordingly, DKD exploits a small frame encoder $\mathrm{F(\cdot)}$ to learn high-level image representations $f_{t+1}$ of frame $I_{t+1}$ to match these distilled pose kernels, alleviating the demand of large networks that troubles prior works~\cite{song2017thin,luo2017lstm}.
Then, DKD applies the distilled pose kernels $k_t$ on feature maps $f_{t+1}$, in a sliding window manner to search for the region with similar patterns as each body joint, namely,
\begin{equation}\label{eq:cross_correlations}
h_{t+1}^j=k_t^j \otimes f_{t+1},
\end{equation}
where $\otimes$ denotes the convolution operation, and $h_{t+1}^j{\in}\mathbb{R}^{m{\times}n}$, $k_t^j{\in}\mathbb{R}^{S{\times}S{\times}C}$ are the confidence map and pose kernels of the $j$th joint, respectively. With the above formulation, DKD casts human pose estimation to a matching problem and locates the position with maximum response on $h_{t+1}^j$ in the $(t{+}1)$th frame as the $j$th body joint. 

In this way, the pose kernel distillator equips DKD with the capability of transferring pose knowledge among neighboring frames and enables  small network to estimate human pose in videos. Its distilled pose kernels can be applied to fast localize body joints with simple convolution, further improving the efficiency.
In addition, it can directly leverage temporal cues of one frame to assist body joint localization in the following frame, without requiring auxiliary optical flow models~\cite{song2017thin} or decoders appended to RNN units~\cite{luo2017lstm}. It can also fast distill pose kernels in a one-shot manner, avoiding complex iterating utilized by previous online kernel learning models~\cite{bertinetto2016fully,valmadre2017end}. 
Moreover, it framewisely updates pose kernels and improves the robustness of our model to joint appearance and configuration variations.

It is worth noting that, for the first frame, due to the lack of preceding temporal cues, we utilize another pose model $\mathrm{P}(\cdot)$, usually larger than $\mathrm{F(\cdot)}$, to initialize its confidence map, \emph{i.e.}, $h_1{=}\mathrm{P}(I_1)$. 
In particular, $\Phi(\cdot)$ together with $\mathrm{F}(\cdot)$ and $\mathrm{P}(\cdot)$ instantiate the \emph{pose generator}. 
Given pose annotations $\{\hat{h}_t\}_{t=1}^T$, to learn the pose generator, we define the loss as
\begin{equation}
\mathcal{L}_{\mathrm{G}}{=}\sum_{t=1}^{T}\ell_2(h_t, \hat{h}_t),
\end{equation}
where $\ell_2$ denotes the Mean Square Error loss.

\vspace{-4mm}
\paragraph{Temporally adversarial training}  To further leverage temporal cues, DKD adopts the adversarial training strategy to learn proper supervision in the temporal dimension for improving the pose kernel distillator. Adversarial training was only exploited for 
images
in the spatial dimension in prior works~\cite{chou2017self,chen2017adversarial}. In contrast, our proposed temporally adversarial training strategy aims to provide constraints for pose changes in the temporal dimension, helping estimate coherent human poses in consecutive frames of videos. Inspired by~\cite{chou2017self}, DKD introduces a discriminator $\mathrm{D}(\cdot)$ to distinguish the changes of groundtruth confidence maps between neighboring frames from predicted ones. The discriminator $\mathrm{D}(\cdot)$ takes as input two neighboring confidence maps (either from groundtruth or prediction) concatenated with the corresponding images, and reconstructs the change of the confidence maps. For real (groundtruth) samples $\hat{h}_t$ and $\hat{h}_{t+1}$, the discriminator $\mathrm{D}(\cdot)$ targets at approaching their change $\hat{d}_t{=}\hat{h}_{t+1}{-}\hat{h}_t$, while for fake (predicted) samples $h_t$ and $h_{t+1}$, keeping the reconstructed change away from $d_t{=}h_{t+1}{-}h_t$. 
Therefore, the discriminator can better differentiate groundtruth change from erroneous predictions.
In this way, the discriminator $\mathrm{D}(\cdot)$ criticizes pixel-wise variations of confidence maps and judges whether joint positions are in rational movements, to encourage the pose kernel distillator to distill suitable pose kernels and ensure consistency of estimated poses between neighboring frames. To train 
the discriminator 
$\mathrm{D}(\cdot)$, we define its loss function as
\begin{equation}\label{eq:discriminator_loss}
\mathcal{L}_{\mathrm{D}}{=}\lambda\sum_{t=1}^{T-1}\ell_2(d_t^{\mathrm{f}},d_t)-\sum_{t=1}^{T-1}\ell_2(d_t^{\mathrm{r}},\hat{d}_t),
\end{equation}
where $d_t^{\mathrm{r}}{=}\mathrm{D}(I_t,\hat{h}_t,I_{t+1}, \hat{h}_{t+1})$ denotes the output from the discriminator for real samples and $d_t^{\mathrm{f}}{=}\mathrm{D}(I_t,h_t,I_{t+1}, h_{t+1})$ denotes the one for fake samples. $\lambda$ is a variable for dynamically balancing the relative learning speed between the pose generator and temporally adversarial discriminator.

The temporally adversarial training conventionally follows a two-player minmax game. Therefore, the final objective function of the DKD model is written as
\begin{equation}\label{eq:total_loss}
\min_{\mathrm{P},\mathrm{F},\Phi}\max_{\mathrm{D}}\mathcal{L}_{\mathrm{G}} + \eta\mathcal{L}_{\mathrm{D}},
\end{equation}
where $\eta$ is a constant for weighting generator loss and discriminator loss, set as 0.1. The training process to optimize the above object function will be illustrated in Section~\ref{sec:train_and_infer}. 

\subsection{Network architecture}

\paragraph{Pose initializer}
For the first frame $I_1$, DKD utilizes a pose initializer $\mathrm{P}(\cdot)$ to directly estimate its confidence maps $h_1$. 
Here, $\mathrm{P}(\cdot)$ exploits the network following~\cite{xiao2018simple}, which achieves outstanding performance with a simple architecture. The network follows a U-shape architecture. It first encodes down-sized feature maps from the input image, and then gradually recovers high-resolution feature maps by appending several deconvolution layers,
as shown in Fig.~\ref{fig:flow_chart} (b). In particular, we use ResNet~\cite{he2016deep} as the backbone and append two deconvolution layers, resulting in a total stride of the network of 8. The other settings follow~\cite{xiao2018simple}.

\vspace{-4mm}
\paragraph{Frame encoder}
DKD utilizes an encoder $\mathrm{F(\cdot)}$ to extract high-level features $f_t$ of  frame $I_t$ to match the pose kernels from the pose kernel distillator. Here, we design $\mathrm{F(\cdot)}$ with the same network architecture as the pose initializer $\mathrm{P}(\cdot)$, with only the last classification layer  removed from $\mathrm{P}(\cdot)$. Note, the backbone of $\mathrm{F(\cdot)}$ is much smaller than $\mathrm{P}(\cdot)$.

\vspace{-4mm}
\paragraph{Pose kernel distillator}
The pose kernel distillator $\Phi(\cdot)$ in DKD takes as input the temporal information, represented by the concatenation of feature maps $f_t$ and confidence maps $h_t$, and distills the pose kernels $k_t$ in a one-shot feed-forward manner. We implement $\Phi(\cdot)$ with a CNN, including three convolution layers followed by BatchNorm and ReLU layers and two pooling layers. Its architecture is shown in Fig.~\ref{fig:flow_chart} (c). This light-weight CNN guarantees the efficiency of $\Phi(\cdot)$. However, it is inefficient and infeasible for $\Phi(\cdot)$ to directly learn all kernels $k_t{\in}\mathbb{R}^{S{\times}S{\times}C{\times}K}$ due to their large scale which  brings high computational complexity and also the risk of overfitting. To avoid these issues, inspired by~\cite{bertinetto2016learning}, DKD exploits $\Phi(\cdot)$ to learn the kernel bases $k_t^\prime$ instead of full size $k_t$ via performing the following factorization:
\begin{equation}
k_t = U \otimes k_t^\prime \otimes_\mathrm{C} V,
\end{equation}
where $\otimes$ is the convolution operation, $\otimes_{\mathrm{C}}$ the channel-wise convolution, and $U{\in}\mathbb{R}^{1{\times}1{\times}C{\times}K}$, $V{\in}\mathbb{R}^{1{\times}1{\times}C{\times}C}$ are coefficients over the kernel bases $k_t^{\prime}{\in}\mathbb{R}^{S{\times}S{\times}C}$. In this way, the size of actual outputs $k_t^\prime$ from the pose kernel distillator is smaller than original $k_t$ by a magnitude, thus enhancing the efficiency of the DKD model.

To generate the confidence maps $h_{t+1}$ of $I_{t+1}$, the calculation between $k_{t}$ and $f_{t+1}$ is implemented with convolution layers. In particular, we first use a $1{\times}1$ convolution parameterized by $V$ on $f_{t+1}$. Then we apply $k_{t}^{\prime}$ in a dynamic convolution layer~\cite{nie2018mula}, which is the same with traditional convolution layer, just replacing the pre-learned static convolution kernels with the dynamically learned ones. Finally, we adopt another $1{\times}1$ convolution with $U$ to produce $h_{t+1}$. To scale the estimation results with the pose kernels, we add a BatchNorm layer in the last to facilitate the training. 

\vspace{-4mm}
\paragraph{Temporally adversarial discriminator}
DKD utilizes the temporally adversarial discriminator $\mathrm{D}(\cdot)$ to enhance the learning process of the pose kernel distillator with confidence map variations as auxiliary temporal supervision. We design $\mathrm{D}(\cdot)$ with the same network backbone as 
the frame encoder $\mathrm{F}(\cdot)$ to balance the learning capability between pose generator and discriminator.

\subsection{Training and inference}\label{sec:train_and_infer}
In this subsection, we will explain the training and inference process of the DKD model for human pose estimation in videos. Specifically, DKD exploits a temporally adversarial training strategy. The discriminator is optimized via maximizing the loss function defined in Eqn.~\eqref{eq:total_loss} for distinguishing the changes of groundtruth confidence maps from estimated ones between neighboring frames. On the other hand, the generator produces a set of confidence maps for consecutive frames in a video and meanwhile fools the discriminator via making the changes of estimated poses approach those of groundtruth ones.
To synchronize the learning speed between generator and discriminator, we follow~\cite{berthelot2017began,chou2017self} to update $\lambda$ in Eqn.~\eqref{eq:discriminator_loss} for each iteration $i$:
\begin{equation}\label{eq:update_lambda}
\lambda_{i+1} = \lambda_i + \gamma\big(\sum_{t=1}^{T-1}\ell_2(d_t^{\mathrm{r}},\hat{d}_t) - \sum_{t=1}^{T-1}\ell_2(d_t^{\mathrm{f}},d_t)\big)
\end{equation}
where $\gamma$ is a hyper-parameter controlling the update rate and set as 0.1. 
$\lambda$ is initialized as 0 and bounded in $[0,1]$. As defined in Eqn.~\eqref{eq:update_lambda}, when the generator successfully fools the discriminator, $\lambda$ will be increased to make the optimizer emphasize improving the discriminator, and vice versa.
The overall training process is illustrated in Algorithm~\ref{alg:temporal_gan}.

During inference, the 
discriminator $\mathrm{D}(\cdot)$ is removed. Given a video, DKD first utilizes the pose initializer $\mathrm{P}(\cdot)$ to estimate the confidence maps $h_1$ of the first frame. Then, $h_1$ is combined with the feature maps $f_1$ from the encoder $\mathrm{F(\cdot)}$ as input to the pose kernel distillator $\Phi(\cdot)$ for distilling the initial pose kernels $k_1$. For the second and subsequent frames, DKD applies the framewisely updated pose kernels $k_{t}$ on the feature maps $f_{t+1}{=}\mathrm{F}(I_{t+1})$ of the posterior frame to estimate the confidence maps $h_{t+1}$. Finally, DKD outputs body joint positions for each frame by localizing the maximum responses on the corresponding confidence maps. The overall inference procedure of DKD is given in Fig.~\ref{fig:flow_chart} (a).

\begin{algorithm}[t!]\footnotesize
	\caption{Training process for our DKD model.}
	\SetKwInOut{Input}{input} \SetKwInOut{Output}{output}
	\Input{video $\{I_t\}_{t=1}^T$, groundtruth $\{\hat{h}_t\}_{t=1}^T$, iteration number $E$}
	\textbf{initialization:}  $\mathcal{L}_D \leftarrow 0$, $\mathcal{L}_G \leftarrow 0$ \\
	\For{iteration $i$, $i{=}1$ to $E$}{
		Forward pose initializer $h_1 \leftarrow \mathrm{P}(I_1)$ \\
		Update loss $\mathcal{L}_{\mathrm{G}} \leftarrow \ell_2(h_1, \hat{h}_1)$ \\ 
		\For{frame $t$, $t{=}1$ to $T$}{
			\If{$t$ equals 1}{
				Encode image representations $f_1\leftarrow\mathrm{F}(I_1)$ \\
			}
			\Else{
				Forward discriminator $d_{t-1}^{\mathrm{r}}{\leftarrow}\mathrm{D}(I_{t-1}, \hat{h}_{t-1}, I_{t}, \hat{h}_{t})$ \\
				Update loss $\mathcal{L}_{D}\leftarrow\mathcal{L}_{D}-\ell_2(d_{t-1}^{\mathrm{r}}, \hat{d}_{t-1})$ \\
				Update pose kernels $k_{t-1}\leftarrow\Phi(f_{t-1}, h_{t-1})$ \\
				Encode image representations $f_t\leftarrow\mathrm{F}(I_t)$ \\
				Estimate confidence map $h_t$ with Eqn.~\eqref{eq:cross_correlations} \\
				Update loss $\mathcal{L}_{\mathrm{G}} \leftarrow \mathcal{L}_{\mathrm{G}}+\ell_2(h_t, \hat{h}_t)$ \\
				Forward discriminator $d_{t-1}^{\mathrm{f}}{\leftarrow}\mathrm{D}(I_{t-1}, h_{t-1}, I_{t}, h_{t})$ \\
				Update loss $\mathcal{L}_{D}\leftarrow\mathcal{L}_{D}+\lambda_i\ell_2(d_{t-1}^{\mathrm{f}}, d_{t-1})$ \\
				Update loss $\mathcal{L}_{\mathrm{G}} \leftarrow \mathcal{L}_{\mathrm{G}}+\eta\ell_2(d_{t-1}^{\mathrm{f}}, d_{t-1})$ \\
			}
		}	
		Update discriminator $\mathrm{D}(\cdot)$ with $-\mathcal{L}_\mathrm{D}$ via backpropagation \\
		Update $\mathrm{P}(\cdot)$, $\Phi(\cdot)$, and $\mathrm{F}(\cdot)$ with $\mathcal{L}_\mathrm{G}$ via backpropagation \\
		Update $\lambda_i$ with Eqn.~\eqref{eq:update_lambda}
	}
	\label{alg:temporal_gan}
\end{algorithm}

\section{Experiments}

\subsection{Experimental setup}

\paragraph{Datasets} 
We evaluate our model on two widely used benchmarks: Penn Action~\cite{zhang2013actemes} and Sub-JHMDB~\cite{jhuang2013towards}. Penn Action dataset is a large-scale unconstrained video dataset. It contains 2,326 video clips,  1,258 for training and 1,068 for testing. Each person in a frame is annotated with 13 body joints, including the coordinates and visibility. Following conventions, evaluations on the Penn Action dataset only consider the visible joints. Sub-JHMDB is another dataset for video based human pose estimation. It provides labels for 15 body joints. Different from Penn Action dataset, it only annotates visible joints for complete bodies. It contains 316 video clips with 11,200 frames in total. The ratio for the number of training and testing videos is roughly 3:1. In addition, it includes three different split schemes. Following previous works~\cite{luo2017lstm,song2017thin}, we separately conduct evaluations on these three splits and report the average precision.

\vspace{-4mm}
\paragraph{Data augmentation} 
For both the Penn Action dataset and Sub-JHMDB dataset, we perform data augmentation following conventional strategies, including random scaling with a factor from $[0.8, 1.4]$, random rotation in $[-40^\circ, 40^\circ]$ and random flipping. The same augmentation setting is applied to all the frames in a training video clip. In addition, each frame is cropped based on the person center on the original image and padded to $256{\times}256$ as input for training.

\vspace{-4mm}	
\paragraph{Implementation} 
For fair comparison with previous works~\cite{luo2017lstm,song2017thin}, we first pre-train the pose initializer and the frame encoder for single-person pose estimation on the MPII~\cite{andriluka14cvpr} dataset. Then, we fine-tune the pre-trained models together with the randomly initialized pose kernel distillator and the temporally adversarial discriminator on Penn Action dataset and Sub-JHMDB dataset for 40 epochs, respectively. In particular, each training sample contains 5 frames, which are consecutively sampled from a video. We set the channel number $C$ of the pose kernels $k_t$ as 256 and the kernel size $S$ as 7. We implement our DKD model with Pytorch~\cite{paszke2017pytorch} and use RMSprop as the optimizer~\cite{rmsprop2012}. We set the initial learning rate as 0.0005 and drop it with a multiplier 0.1 at the 15th and 25th epochs. For evaluation, we perform seven-scale testing with flipping. 

\vspace{-4mm}	
\paragraph{Evaluation metrics} 
We evaluate the performance  with PCK~\cite{yang2013articulated}\textemdash the localization of a body joint is considered to be correct if it falls within $\alpha{\cdot}L$ pixels of the groundtruth. $\alpha$ controls the relative threshold and conventionally set as 0.2. $L$ is the reference distance, set as $L{=}\max(H,W)$ following prior works~\cite{luo2017lstm,song2017thin} with $H$ and $W$ being  height and width of the person bounding box. We term this metric as PCK normalized by person size. This metric is somewhat loose to  precisely evaluate the model performance as person size is usually relatively large. Thereby, we follow the conventions of still-image based pose estimation~\cite{Johnson11,chu2016structured,yang2016end,wei2016convolutional}, and also adopt another metric that takes torso size as reference distance. We term it as PCK normalized by torso size.

\begin{table}[t!]\scriptsize
	\caption{Ablation studies on Penn Action dataset with PCK normalized by torso size as evaluation metric.}
	\label{tab:ablation_component}
	\centering
	\setlength{\tabcolsep}{1.5pt}
	\begin{tabular}{l>{\columncolor[gray]{0.9}}c|ccccccc>{\columncolor[gray]{0.9}}c}
		\toprule
		Methods & Flops(G) &Head & Sho. & Elb. & Wri. & Hip & Knee  & Ank. & PCK \\
		\midrule
		Baseline(ResNet101) &11.02 & 96.1 & 90.7 & 91.4 & 89.5 & 86.2 & 92.2 & 88.9 & 90.7 \\
		\midrule
		DKD(ResNet50) & 8.65 & 96.6	& 93.7 & 92.9 & 91.2 & 88.8 & 94.3 & 93.7 & 92.9  \\
		DKD(ResNet50)-w/o-TAT  & 8.65   & 96.6 & 92.6 & 92.9 & 90.8 & 87.5 & 93.4 & 92.4 & 92.1  \\
		DKD(ResNet50)-w/o-PKD & 7.66  & 96.0   & 91.8   & 92.4   & 90.4   & 88.3   & 93.5   & 89.8   & 91.6  \\
		Baseline(ResNet50)  & 7.66 & 96.0   & 90.5   & 89.4   & 87.6   & 83.8   & 89.7   & 86.0   & 88.8 \\
		\midrule
		DKD(ResNet34) & 7.68 & 96.4	& 91.9	& 93.0	& 90.8	& 88.6	& 93.5 & 91.9 & 92.1  \\
		DKD(ResNet34)-w/o-TAT & 7.68  & 96.4 &	91.2 & 92.7	& 89.9	& 87.3	& 93.3	& 90.9 & 91.4 \\
		DKD(ResNet34)-w/o-PKD & 6.69  & 95.9   & 91.1   & 91.9   & 89.3   & 87.7   & 92.5   & 90.3   & 91.0  \\
		Baseline(ResNet34) & 6.69 & 95.8   & 88.7   & 88.5   & 86.7   & 83.6   & 89.6   & 85.3   & 87.3  \\
		\midrule
		DKD(ResNet18)  & 5.27 & 95.7	& 90.0 & 92.2 & 89.4 & 86.8 & 92.3 & 89.5 &	90.6  \\
		DKD(ResNet18)-w/o-TAT & 5.27      & 95.5 & 89.3 & 91.9 & 89.1 & 85.0 & 91.6 & 89.0 & 89.9 \\
		DKD(ResNet18)-w/o-PKD & 4.28 & 95.0   & 89.1   & 92.4   & 88.7   & 85.5   & 91.4   & 87.7   & 89.7  \\
		Baseline(ResNet18) & 4.28  & 94.7 & 86.0 & 87.7 & 84.6 & 81.1 & 87.4 & 84.3 & 86.1 \\
		\bottomrule
	\end{tabular}
	\vspace{-3mm}
\end{table}

\begin{figure}[t!]
	\begin{center}
		\includegraphics[scale=0.52]{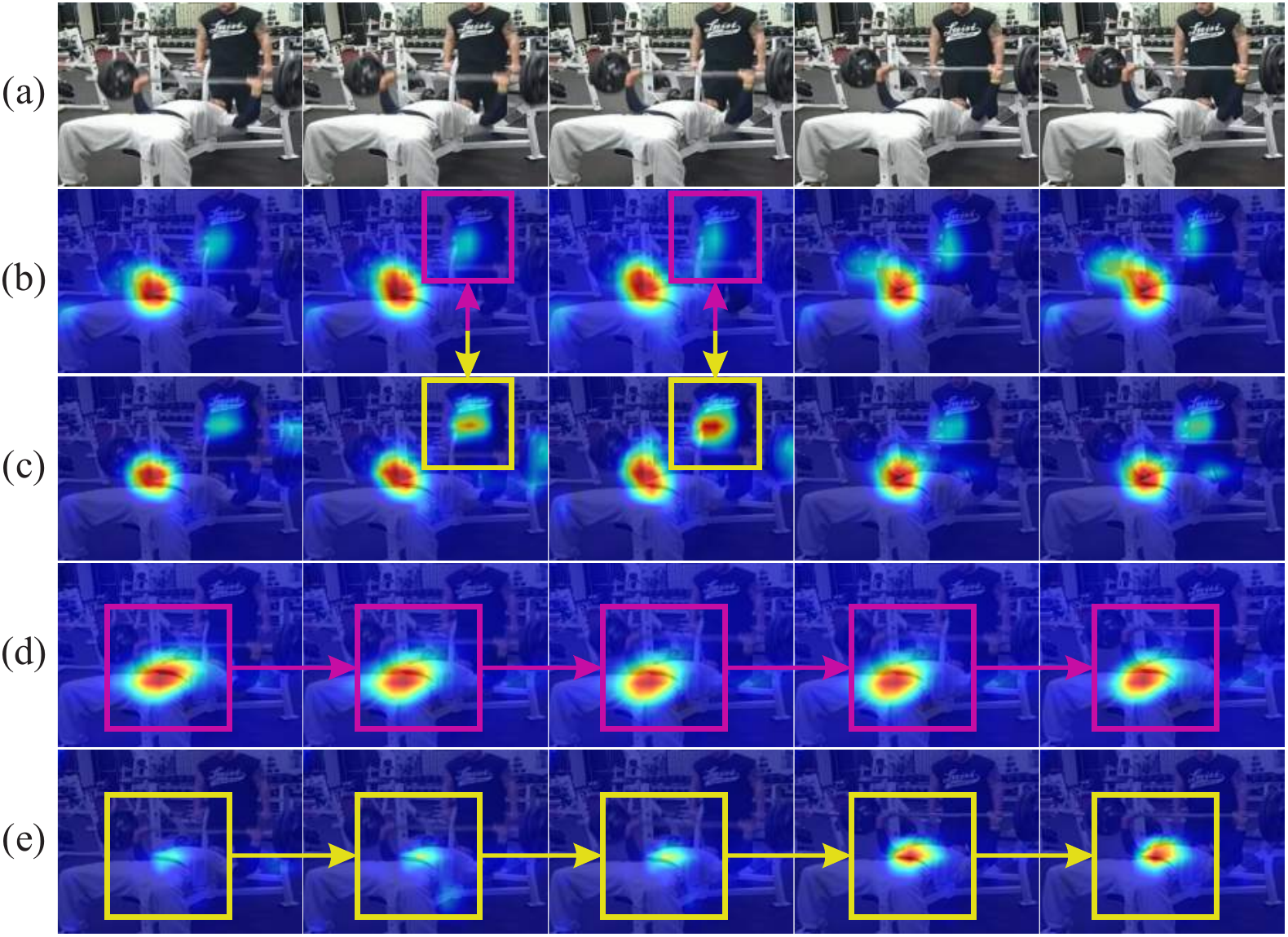}
		\caption{Comparison of confidence maps estimated from the proposed model DKD(ResNet34) and the baseline one Baseline(ResNet34). (a) are input frames. (b) and (d) are estimated confidence maps from our model for right elbow and right hip, respectively, and (c) and (e) from baseline. Best viewed in color.}
		\label{fig:comp_with_baseline}
	\end{center}
	\vspace{-8mm}
\end{figure}	

\subsection{Ablation analysis}

We first conduct ablation studies on Penn Action dataset to analyze the efficacy of each core component of our DKD model: the pose kernel distillator and the temporally adversarial training. We fix the backbone of the pose initializer as ResNet101. We vary the backbone of  frame encoder ranging in ResNet18/34/50, since it dominates 
the computational cost of pose estimation of our model. We use DKD(ResNet$x$) to denote our full model, where $x{\in}\{18,34,50\}$ represents the backbone  depth of  the frame encoder. We use DKD(ResNet$x$)-w/o-TAT to denote the model without the temporally adversarial training and DKD(ResNet$x$)-w/o-PKD the model without the pose kernel distillator. We use Baseline(ResNet$x$) to denote the single-image pose estimation model without using temporal cues. Results are shown in Tab.~\ref{tab:ablation_component}. 

From Tab.~\ref{tab:ablation_component}, we can see that DKD(ResNet34) and DKD(ResNet50) use smaller networks for frame feature learning while achieve much better performance than Baseline(ResNet101) which is much deeper. We can also see DKD(ResNet18) achieves comparable performance to Baseline(ResNet101) ($90.6\%$ PCK vs $90.7\%$ PCK), with up to $2{\times}$ flop reduction (5.27G vs 11.02G Flops). These results verify the efficacy of DKD to enable  small networks to estimate human pose in videos, bring efficiency enhancement while achieving outperforming accuracy. 

By comparing the DKD(ResNet$x$)-w/o-TATs and the Baseline(ResNet$x$)s, we find that the computation overhead of the pose kernel distillator is small, only bringing slight flops increase,
\emph{e.g.}, with ResNet50 as backbone, from 7.66G to 8.65G. 
We can also find the pose kernel distillator improves frame-level performance for human pose estimation over baselines by $4.3\%$ in average.  Besides,  DKD(ResNet$x$)-w/o-TATs always outperform DKD(ResNet$x$)-w/o-PKDs, 
this implies the distilled pose kernels carry  knowledge of body joint patterns and provide compact guidance for pose estimation between neighboring frames, which are absent in still-image based inference.
The above results verify the efficacy of the pose kernel distillator for efficiently transferring pose knowledge to assist poses estimation in videos. 

By comparing the time cost of the DKD(ResNet$x$)-w/o-PKDs and the Baseline(ResNet$x$)s, we find temporally adversarial training does not hurt inference speed, since the discriminator is used only in training. In addition, the temporally adversarial training consistently improves the baseline performance for all body joints, in particular for the joints difficult to localize,
\emph{e.g.}, DKD(ResNet34)-w/o-PKD improves the accuracy of ankles from $85.3\%$ PCK to $90.3\%$ PCK. This demonstrates the proposed temporally adversarial training is effective for regularizing temporal changes over pose predictions during model training. 

Combining temporally adversarial training with the pose kernel distillator, the full DKD model further boosts the performance over all the ablated models, showing they are complementary to each other. Especially, DKD(ResNet$x$)s achieves average $5.5\%$ performance gain over the corresponding vanilla baselines Baseline(ResNet$x$)s. 

To better reveal the  advantages of our DKD model over single-frame based models, we visualize the confidence maps estimated from DKD(ResNet34) and Baseline (ResNet34) for the elbow and ankle in Fig.~\ref{fig:comp_with_baseline}. By comparing Fig.~\ref{fig:comp_with_baseline} (b) and (c), we can observe that our DKD model produces pose kernels of the correct person of interest with more accurate response. In contrast,  the baseline model produces false alarms on the elbow of another person in the frame. We can also see that the proposed model can produce consistent confidence maps for the hip in Fig.~\ref{fig:comp_with_baseline} (d) while the baseline model produces unstable estimations even with  fixed hip  Fig.~\ref{fig:comp_with_baseline} (e). These results further validate the capability of the proposed model for generating accurate and temporally consistent human pose estimations  in videos.

\begin{table}[t!]\scriptsize
	\caption{Comparison of temporally \emph{vs.} spatially adversarial training, and pose kernel distillator \emph{vs.} Convolutional LSTM. The accuracy is measured with PCK normalized by torso size.}
	\label{tab:ablation_jtl_and_tal}
	\centering
	\setlength{\tabcolsep}{1.5pt}
	\begin{tabular}{l>{\columncolor[gray]{0.9}}c|ccccccc>{\columncolor[gray]{0.9}}c}
		\toprule
		Methods & Flops(G) &Head & Sho. & Elb. & Wri. & Hip & Knee  & Ank. & PCK \\
		\midrule
		DKD(ResNet34) & 7.68 & 96.4	& 91.9	& 93.0	& 90.8	& 88.6	& 93.5 & 91.9 & 92.1 \\
		DKD(ResNet34)-w-SAT & 7.68 & 96.4 &	91.4 & 92.8	& 90.1	& 87.7	& 93.4	& 91.2 & 91.6 \\
		DKD(ResNet34)-w-LSTM & 10.16 & 95.7 & 89.5 & 92.9 & 90.2 & 86.9 & 93.5 & 90.1 & 91.1 \\
		\bottomrule
	\end{tabular}
	\vspace{-4mm}
\end{table}

Next, we  analyze how well  our pose kernel distillator performs for propagating temporal information  via comparing it with the state-of-the-art Convolutional LSTMs~\cite{luo2017lstm}. We also compare our temporally adversarial training with the spatially one in~\cite{chou2017self}.
All the compared models adopt the ResNet101 as the backbone of the pose initializer and ResNet34 as the frame encoder. Except for the compared components, all the other settings are the same. Results are shown in Tab.~\ref{tab:ablation_jtl_and_tal}. We use DKD(ResNet34)-w-LSTM to denote the model utilizing Convolutional LSTM for temporal cues propagation instead of our pose kernel distillator in the DKD model. We can observe that DKD(ResNet34)-w-LSTM degrades the accuracy of DKD(ResNet34) for all body joints, especially for wrist and ankle. In addition, it increases the flops from 7.68G to 10.16G. These results evaluate the superiority of the pose kernel distillator in both efficiency and efficacy for transferring pose knowledge between neighboring  frames over traditional RNN units.

We use DKD(ResNet34)-w-SAT to denote the model in which our temporally adversarial training is replaced with the spatially one in~\cite{chou2017self}. Specifically, \cite{chou2017self} introduces a discriminator to distinguish the single-frame groundtruth confidence maps from estimated ones for obtaining structural spatial constraints on poses. We can see DKD(ResNet34) consistently outperforms DKD(ResNet34)-w-SAT. In addition, by comparing DKD(ResNet34)-w-SAT with DKD(ResNet34)-w/o-TAT in Tab.~\ref{tab:ablation_component},  spatially adversarial training only brings limited improvement. These results further verify the efficacy of using adversarial training in temporal dimension. 

\subsection{Comparisons with state-of-the-arts}

Tab.~\ref{tab:cmp_sota_penn} show the comparisons of our DKD model with state-of-the-arts on Penn Action dataset. In particular, the method proposed in~\cite{luo2017lstm} follows the Encoder-RNNs-Decoder framework with Convolutional LSTMs, while~\cite{song2017thin} exploits optical flow models to align confidence maps of neighboring frames. We report the performance of our model with both person and torso size as reference distance under the PCK evaluation metric. For comparison with current best model~\cite{luo2017lstm}, we report both its performance with PCK normalized by torso size, flops and running time\footnote{We reproduce the results of~\cite{luo2017lstm} with PCK normalized by torso size via running the codes released by the authors on the repo: https://github.com/lawy623/LSTM\_Pose\_Machines. The running time is counted on GPU GTX 1080ti for both~\cite{luo2017lstm} and our model.}. For our DKD model, we fix the backbone of the pose initializer as ResNet101. We vary the backbone of  frame encoder ranging in ResNet18/34/50. Since both of state-of-the-arts~\cite{luo2017lstm} and~\cite{song2017thin} use the same network as Convolutional Pose Machines (CPM)~\cite{wei2016convolutional}, we also experiment our DKD model with a frame encoder as a simplified version of CPM by replacing its kernels with size larger than 3 to $3{\times}3$ kernels, denoted as DKD(SmallCPM), to further verifying the efficacy of DKD to facilitate small networks in video-based pose estimation. 

\begin{table}\scriptsize
	\caption{Comparison with state-of-the-arts on Penn Action dataset.} 
	\label{tab:cmp_sota_penn}
	\centering
	\setlength{\tabcolsep}{1pt}
	\begin{tabular}{l>{\columncolor[gray]{0.9}}c>{\columncolor[gray]{0.9}}c|ccccccc>{\columncolor[gray]{0.9}}c}
		\toprule
		Methods & Flops(G) & Time(ms) &Head & Sho. & Elb. & Wri. & Hip & Knee  & Ank. & PCK \\
		\midrule
		\multicolumn{3}{c|}{ } & \multicolumn{8}{c}{Normalized by Person Size} \\
		Park \emph{et al.}~\cite{park2011n} & - & -  & 62.8 & 52.0 & 32.3 & 23.3 & 53.3 & 50.2& 43.0 & 45.3 \\
		Nie \emph{et al.}~\cite{xiaohan2015joint} & - & -  & 64.2 & 55.4 & 33.8 & 24.4 & 56.4 & 54.1 & 48.0 & 48.0 \\
		Iqal \emph{et al.}~\cite{iqbal2017pose} & - & -  & 89.1 & 86.4 & 73.9 & 73.0 & 85.3 & 79.9 & 80.3 & 81.1 \\
		Gkioxari \emph{et al.}~\cite{gkioxari2016chained} & - & -   & 95.6 & 93.8 & 90.4 & 90.7 & 91.8 & 90.8 & 91.5 & 91.8 \\
		Song \emph{et al.}~\cite{song2017thin} & - & -  & 98.0  & 97.3 & 95.1 & 94.7 & 97.1 & 97.1 & 96.9 & 96.5 \\
		Luo \emph{et al.}~\cite{luo2017lstm} &  70.98 & 25   & \textbf{98.9} & 98.6 & 96.6 & 96.6 & 98.2  & \textbf{98.2} & \textbf{97.5} & 97.7 \\
		DKD(SmallCPM) & 9.96 & 12 & 98.4 & 97.3 & 96.1 & 95.5 & 97.0 & 97.3 & 96.6  & 96.8\\
		DKD(ResNet50) & \textbf{8.65} & \textbf{11}  & 98.8   & \textbf{98.7}   & \textbf{96.8}   & \textbf{97.0}  & \textbf{98.2}   & 98.1  & 97.2   & \textbf{97.8} \\
		\midrule
		\multicolumn{3}{c|}{ } & \multicolumn{8}{c}{Normalized by Torso Size} \\
		Luo \emph{et al.}~\cite{luo2017lstm} & 70.98 & 25  & 96.0 & 93.6 & 92.4 & 91.1 & 88.3 & 94.2 & 93.5 & 92.6 \\
		DKD(SmallCPM) & 9.96 & 12 & 96.0 & 93.5 & 92.0 & 90.6 & 87.8 & 94.0 & 93.1 & 92.4\\
		DKD(ResNet50) & \textbf{8.65} & \textbf{11} & \textbf{96.6} & \textbf{93.7} & \textbf{92.9} & \textbf{91.2} & \textbf{88.8} & \textbf{94.3} & \textbf{93.7} & \textbf{92.9} \\
		\bottomrule
	\end{tabular}
	\vspace{-2mm}
\end{table}

\begin{figure}[t!]
	\begin{center}
		\includegraphics[scale=0.45]{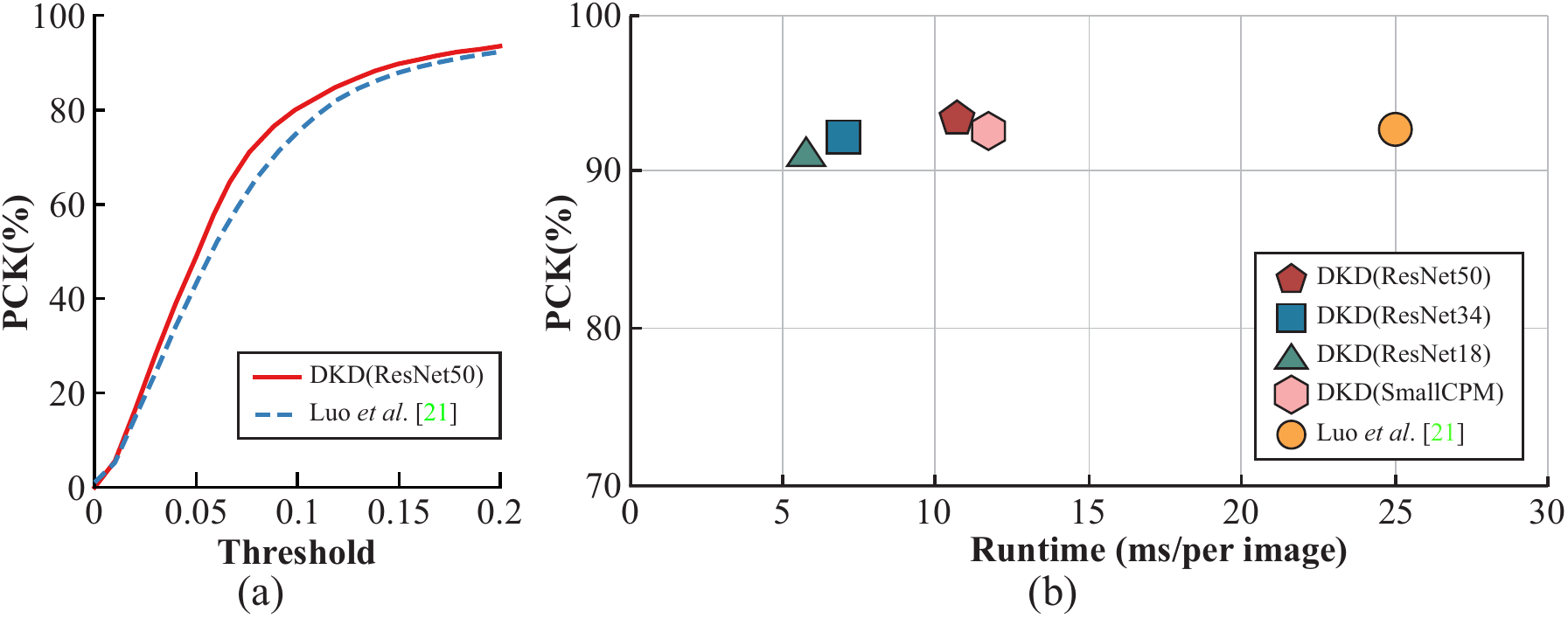}
		\captionof{figure}{Extensive analysis for comparing our method with state-of-the-art~\cite{luo2017lstm} on (a) PCK over different thresholds with $\alpha$ ranging from 0 to 0.2; (b) speed \emph{vs.} accuracy.}
		\label{fig:var_alpha}
	\end{center}
	\vspace{-8mm}
\end{figure}

\begin{figure*}[t!]
	\begin{center}
		\includegraphics[scale=0.61]{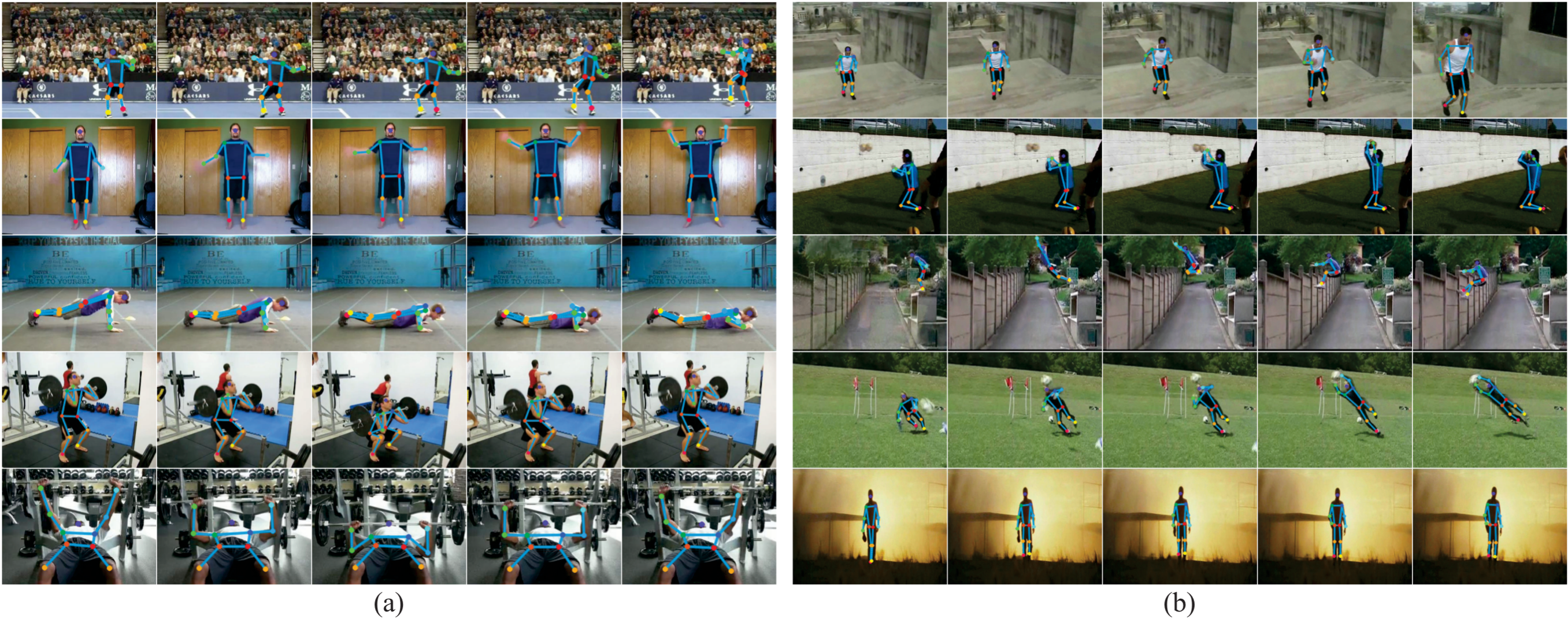}
		\vspace{-1mm}
		\caption{Qualitative results on (a) Penn Action dataset and (b) Sub-JHMDB dataset. Best viewed in color and 2x zoom.}
		\label{fig:vis_examples}
	\end{center}
	\vspace{-8mm}
\end{figure*} 

From Tab.~\ref{tab:cmp_sota_penn}, we can observe that our best model DKD(ResNet50) reduces the computation flops by a magnitude over~\cite{luo2017lstm} (8.65G vs 70.98G) and achieves 2x faster speed (11ms vs 25ms per image), verifying the outperforming efficiency of our model. 
In addition, we can see under PCK  normalized by person size, DKD(ResNet50) achieves comparable accuracy with state-of-the-art~\cite{luo2017lstm}. When using PCK normalized by torso size, DKD(ResNet50) achieves superior accuracy over~\cite{luo2017lstm}  ($92.9\%$ PCK vs $92.6\%$ PCK) and with better performance for all of the body joints. We also compare our model with~\cite{luo2017lstm} via evaluating the performance with PCK normalized by torso size when varying threshold $\alpha$ from 0 to 0.2 with 0.01 as the step size, and results are shown in Fig.~\ref{fig:var_alpha} (a). We can see that DKD consistently outperforms~\cite{luo2017lstm} under more critic metrics by decreasing $\alpha$. These results demonstrate the superior speed and accuracy of our model for human pose estimation in videos. 

By comparing DKD(SmallCPM) with~\cite{luo2017lstm}, we can find our DKD model maintains high accuracy ($92.4\%$ PCK vs $92.6\%$ PCK) in case of significant simplification to the network (9.96G vs 70.98G Flops).
This result verifies the effectiveness of our DKD model for alleviating the demands of large networks for video-based human pose estimation.

To evaluate the effects of different frame encoder backbones on the efficiency and efficacy of DKD, we plot speed \emph{vs.} accuracy analysis for different models in Fig.~\ref{fig:var_alpha} (b). We can observe that reducing depth of frame encoder backbone from ResNet50 to ResNet18 slightly degrades the accuracy, but speeds up 2x from 11ms to 6.5ms per image. In addition, we can see that DKD(ResNet18) achieves comparable performance with~\cite{luo2017lstm} but 4x faster. These results further validate the efficacy of our DKD model to facilitate small networks in video-based pose estimation. 

\begin{table}[t!]\scriptsize
	\caption{Comparison with state-of-the-arts on Sub-JHMDB dataset.}
	\label{tab:cmp_sota_jhmdb}
	\centering
	\setlength{\tabcolsep}{4.5pt}
	\begin{tabular}{lccccccc>{\columncolor[gray]{0.9}}c}
		\toprule
		Methods &Head & Sho. & Elb. & Wri. & Hip & Knee  & Ank. & PCK\\
		\midrule
		\multicolumn{9}{c}{Normalized by Person Size} \\
		Park \emph{et al.}~\cite{park2011n}   & 79.0 & 60.3 & 28.7 & 16.0 & 74.8 & 59.2 & 49.3 & 52.5 \\
		Nie \emph{et al.}~\cite{xiaohan2015joint}  & 80.3 & 63.5 & 32.5 & 21.6 & 76.3 & 62.7 & 53.1 & 55.7  \\
		Iqal \emph{et al.}~\cite{iqbal2017pose}  & 90.3 & 76.9 & 59.3 & 55.0 & 85.9 & 76.4 & 73.0 & 73.8 \\
		Song \emph{et al.}~\cite{song2017thin} & 97.1 & 95.7 & 87.5 & 81.6 & 98.0 & 92.7 & 89.8 & 92.1 \\
		Luo \emph{et al.}~\cite{luo2017lstm} & 98.2 & 96.5 & 89.6 & 86.0 & 98.7 & 95.6 & 90.9 & 93.6 \\
		DKD(ResNet50) & \textbf{98.3}	& \textbf{96.6}	& \textbf{90.4}	& \textbf{87.1}	& \textbf{99.1}	& \textbf{96.0}	& \textbf{92.9}	& \textbf{94.0} \\
		\midrule
		\multicolumn{9}{c}{Normalized by Torso Size} \\
		Luo \emph{et al.}~\cite{luo2017lstm}   & 92.7 & 75.6 & 66.8 & 64.8 & 78.0 & 73.1 & 73.3 & 73.6 \\
		DKD(ResNet50) & \textbf{94.4} & \textbf{78.9} & \textbf{69.8} & \textbf{67.6} & \textbf{81.8} & \textbf{79.0} & \textbf{78.8} & \textbf{77.4} \\
		\bottomrule
	\end{tabular}
	\vspace{-3mm}
\end{table}

Tab.~\ref{tab:cmp_sota_jhmdb} show the comparisons of our DKD model with state-of-the-arts on Sub-JHMDB dataset. We can see that our DKD model achieves new state-of-the-art $94.0\%$ PCK and performs best for all the body joints. When using the stricter metric PCK normalized by torso size, the superiority of our model over~\cite{luo2017lstm} is more significant, achieving  over $5\%$ improvement ($77.4\%$ PCK vs $73.6\%$ PCK) on average. In addition, we can find that our model  well applies to
small-scale datasets, such as Sub-JHMDB with only 316 videos. These small datasets are challenging since they provide only limited training samples, while in our DKD model, the one-shot pose kernel distillator is able to fast adapt pose kernels, without requiring a large number of training samples for iteratively tuning classifiers as in existing methods.

\vspace{-4mm}
\paragraph{Qualitative results}
Fig.~\ref{fig:vis_examples} shows the qualitative results to visualize  efficacy of the DKD model for human pose estimation in videos on Penn Action and Sub-JHMDB, respectively. We can observe  DKD can accurately estimate human poses in various challenging scenarios, \emph{e.g.}, cluttered backgrounds (the 1st row of Fig.~\ref{fig:vis_examples} (a)), scale variations (the 1st row of Fig.~\ref{fig:vis_examples} (b)), motion blur (the 2nd rows of Fig.~\ref{fig:vis_examples} (a) and (b)). In addition, it can leverage temporal cues to handle occasional disappearance of a body joint caused by occlusion, as shown in the 3rd row of Fig.~\ref{fig:vis_examples} (a), and encourage pose consistency in presence of fast and large-degree pose variations, as shown in the 3rd and 4th rows of Fig.~\ref{fig:vis_examples} (b). Moreover, it is robust to various view-point and lighting conditions, as shown in the 5th rows of Fig.~\ref{fig:vis_examples} (a) and (b), respectively. These results further verify the effectiveness of DKD. 

\section{Conclusion}
This paper presents a Dynamic Kernel Distillation (DKD) model for improving efficiency of human pose estimation in videos. In particular, it adopts a pose kernel distillator to online distill the pose kernels from temporal cues of one frame in a one-shot feed-forward manner. The distilled pose kernels encode knowledge of body joint patterns and provide compact guidance for pose estimation in the posterior frame. With these pose kernels, DKD simplifies body joint localization into a matching procedure with simple convolution. In this way, DKD fast transfers pose knowledge between neighboring frames and enables small networks to accurately estimate human poses in videos, thus significantly lifting the efficiency.
DKD also introduces the temporally adversarial training strategy via constraining the changes of estimated confidence maps between neighboring frames. The whole framework can be end-to-end trained and inferred. Experiments on two benchmarks demonstrate that our model achieves state-of-the-art efficiency with only 1/10 flops and 2x faster speed of the previous best model, and also outperforming accuracy for human pose estimation in videos.

\section*{Acknowledgement}
Jiashi Feng was partially supported by NUS IDS R-263-000-C67-646,  ECRA R-263-000-C87-133 and MOE Tier-II R-263-000-D17-112.

{\small
	\bibliographystyle{ieee_fullname}
	\bibliography{adel}
}

\end{document}